\documentclass[times,art10,twocolumn,latex8]{article}

\usepackage[utf8]{inputenc}
\usepackage{cvpr}
\usepackage{times}
\usepackage{epsfig}
\usepackage{graphicx}
\usepackage{amsmath}
\usepackage{amssymb}

\usepackage{wrapfig}
\usepackage[breaklinks=true,bookmarks=false]{hyperref}
\usepackage{fancyvrb}

\newcommand{\RR}[0]{\mathbb{R}}

\cvprfinalcopy 

\def\cvprPaperID{22} 

\ifcvprfinal\fi

\graphicspath{{./figures/}}

\begin{document}

\title{ReSeg: A Recurrent Neural Network-based Model \\for Semantic Segmentation}

\author{Francesco Visin\thanks{%
Dipartimento di Elettronica
Informazione e Bioingegneria,
Politecnico di Milano,
Milan, 20133, Italy}
~\thanks{%
Montreal Institute for Learning Algorithms (MILA),
University of Montreal,
Montreal, QC, H3T 1J4, Canada
}\\
{\tt\small francesco.visin@polimi.it}
\\
Adriana Romero\footnotemark[2]\\
{\tt\small adriana.romero.soriano@umontreal.ca}
\and
Marco Ciccone\footnotemark[1]\\
{\tt\small marco.ciccone@mail.polimi.it}
\\
Kyle Kastner\footnotemark[2]\\
{\tt\small kyle.kastner@umontreal.ca}
\and
\hspace{1.2cm}
Kyunghyun Cho\thanks{%
Courant Institute and Center for Data Science,
New York University,
New York, NY 10012, United States}\\
\hspace{1.2cm}
{\tt\small kyunghyun.cho@nyu.edu}
\\
\hspace{1.2cm}
Matteo Matteucci\footnotemark[1]\\
\hspace{1.2cm}
{\tt\small matteo.matteucci@polimi.it}
\and
\hspace{1cm}
Yoshua Bengio\footnotemark[2]~~\thanks{%
CIFAR Senior Fellow}\\
\hspace{1cm}{\tt\small yoshua.bengio@umontreal.ca}
\\
\hspace{1cm}
Aaron Courville\footnotemark[2]\\
\hspace{1cm}
{\tt\small aaron.courville@umontreal.ca}
}

\maketitle

\begin{abstract}
We propose a structured prediction architecture, which exploits the local
generic features extracted by Convolutional Neural Networks and the capacity of
Recurrent Neural Networks (RNN) to retrieve distant dependencies. The proposed
architecture, called ReSeg, is based on the recently introduced ReNet model for
image classification. We modify and extend it to perform the more challenging
task of semantic segmentation. Each ReNet layer is composed of four RNN that
sweep the image horizontally and vertically in both directions, encoding
patches or activations, and providing relevant global information. Moreover,
ReNet layers are stacked on top of pre-trained convolutional layers, benefiting
from generic local features. Upsampling layers follow ReNet layers to recover
the original image resolution in the final predictions. The proposed ReSeg
architecture is efficient, flexible and suitable for a variety of semantic
segmentation tasks. We evaluate ReSeg on several widely-used semantic
segmentation datasets: Weizmann~Horse, Oxford~Flower, and CamVid; achieving
state-of-the-art performance. Results show that ReSeg can act as a suitable
architecture for semantic segmentation tasks, and may have further applications
in other structured prediction problems. The source code and model
hyperparameters are available on
\href{https://github.com/fvisin/reseg}{https://github.com/fvisin/reseg}.
\end{abstract}

\section{Introduction}

In recent years, Convolutional Neural Networks (CNN) have become the {\em de
facto} standard in many computer vision tasks, such as image classification and
object detection \cite{Krizhevsky-2012,Erhan2014}. Top performing image
classification architectures usually involve {\em very} deep CNN trained in a
supervised fashion on a large datasets
\cite{Lin2014,Simonyan2015,szegedy2014going} and have been shown to
produce generic hierarchical visual representations that perform well on a
wide variety of vision tasks. However, these deep CNNs heavily reduce the
input resolution through successive applications of
pooling or subsampling layers. While these layers seem to contribute
significantly to the desirable invariance properties of deep CNNs, they
also make it challenging to use these pre-trained CNNs for tasks such as
semantic segmentation, where a per pixel prediction is required.

Recent advances in semantic segmentation tend to convert the standard deep
CNN classifier into Fully Convolutional Networks
(FCN)~\cite{long2014fully,noh2015learning,badrinarayanan2015segnet,
Ronneberger2015} to obtain coarse image representations, which are subsequently
upsampled to recover the lost resolution. However, these methods are not
designed to take into account and preserve both \emph{local} and \emph{global} contextual dependencies,
which has shown to be useful for semantic segmentation
tasks~\cite{Singh2013,Gatta14-deepvision}.
These models often employ Conditional Random Fields (CRFs) as a
post-processing step to locally smooth the model predictions, however the
long-range contextual dependencies remain relatively unexploited.

Recurrent Neural Networks (RNN) have been introduced in the literature to
retrieve global spatial dependencies and further improve semantic
segmentation~\cite{Pinheiro:2014, Gatta14-deepvision, chen2015semantic,
byeon2015scene}. However, training spatially recurrent neural networks tends to
be computationally intensive.

In this paper, we aim at the {\em efficient} application of Recurrent Neural
Networks RNN to retrieve contextual information from images. We propose to
extend the ReNet architecture~\cite{visin2015renet}, originally designed for
image classification, to deal with the more ambitious task of semantic
segmentation. ReNet layers can efficiently capture contextual dependencies from
images by first sweeping the image horizontally, and then sweeping the output
of hidden states vertically.  The output of a ReNet layer is therefore
implicitly encoding the local features at each pixel position with respect to
the whole input image, providing relevant global information. Moreover, in
order to {\em fully} exploit local and global pixel dependencies, we stack the
ReNet layers on top of the output of a FCN, i.e. the intermediate convolutional
output of VGG-16~\cite{Simonyan2015}, to benefit from generic local features.
We validate our method on Weizmann~Horse and Oxford~Flower
foreground/background segmentation datasets as a proof of concept for the
proposed architecture.  Then, we evaluate the performance in the standard
benchmark of urban scenes CamVid; achieving state-of-the-art in all three
datasets.~\footnote{Subsequent but independent
work~\cite{DBLP:journals/corr/YanZJBY16} investigated the combination of ReSeg
with Fully Convolutional Network (FCN) and CRFs, reporting state of the art
results on Pascal VOC.}


\section{Related Work}

Methods based on FCN tackle the information recovery (upsampling) problem in a
large variety of ways. For instance, Eigen et al.~\cite{Eigen2015} introduce a
multi-scale architecture, which extracts coarse predictions, which are then
refined using finer scales. Farabet et al.~\cite{Farabet:2013} introduce a
multi-scale CNN architecture; Hariharan et al.~\cite{Hariharan2015} combine the
information distributed over all layers to make accurate predictions. Other
methods such as~\cite{long2014fully,badrinarayanan2015segnet} use simple
bilinear interpolation to upsample the feature maps of increasingly abstract
layers. More sophisticated upsampling methods, such as
unpooling~\cite{badrinarayanan2015segnet,noh2015learning} or
deconvolution~\cite{long2014fully}, are introduced in the literature. Finally,
\cite{Ronneberger2015} concatenate the feature maps of the downsampling layers
with the feature maps of the upsampling layers to help recover finer
information.

RNN and RNN-like models have become increasingly popular in the semantic
segmentation literature to capture long distance pixel
dependencies~\cite{Pinheiro:2014, Gatta14-deepvision,
byeon2015scene,stollenga2015parallel}. For instance, in~\cite{Pinheiro:2014,
Gatta14-deepvision}, CNN are unrolled through different time steps to include
semantic feedback connections. In~\cite{byeon2015scene}, 2-dimensional Long
Short Term Memory (LSTM), which consist of 4 LSTM blocks
scanning all directions of an image (left-bottom, left-top, right-top, right-bottom),
are introduced to learn long range spatial dependencies. Following a similar direction,
in~\cite{stollenga2015parallel}, multi-dimensional LSTM are swept along
different image directions; however, in this case, computations are re-arranged
in a pyramidal fashion for efficiency reasons. Finally,
in~\cite{visin2015renet}, ReNet is proposed to model pixel dependencies in the
context of image classification. It is worth noting that one important
consequence of the adoption of the ReNet spatial sequences is that they are
even more easily parallelizable, as each RNN is dependent only along a horizontal or
vertical sequence of pixels; i.e., all rows/columns of pixels can be processed
at the same time.

%

\section{Model Description}

The proposed ReSeg model builds on top of ReNet~\cite{visin2015renet} and
extends it to address the task of semantic segmentation. The model
pipeline involves multiple stages.

First, the input image is processed with the first layers of
VGG-16~\cite{Simonyan2015} network, pre-trained on
ImageNet~\cite{imagenet_cvpr09} and not fine-tuned, and is set such that the
image resolution does not become too small. The resulting feature maps are then
fed into one or more \emph{ReNet layers} that sweep over the image. Finally,
one or more \emph{upsampling layers} are employed to resize the last feature
maps to the same resolution as the input and a softmax non-linearity is applied
to predict the probability distribution over the classes for each pixel.

The recurrent layer is the core of our architecture and is composed by multiple
RNN that can be implemented as a vanilla $\tanh$ RNN layer, a Gated Recurrent
Unit~(GRU)~layer~\cite{Cho2014} or a
LSTM~layer~\cite{Hochreiter+Schmidhuber-1997}. Previous work has shown that
the ReNet model can perform well with little concern for the specific recurrent
unit used, therefore, we have chosen to use GRU units as they strike a good
balance between memory usage and computational power.

In the following section we will define the recurrent and the upsampling layers
in more detail.

\subsection{Recurrent layer}
\begin{figure}[t]
    \begin{center}
        \includegraphics[width=0.3\columnwidth]{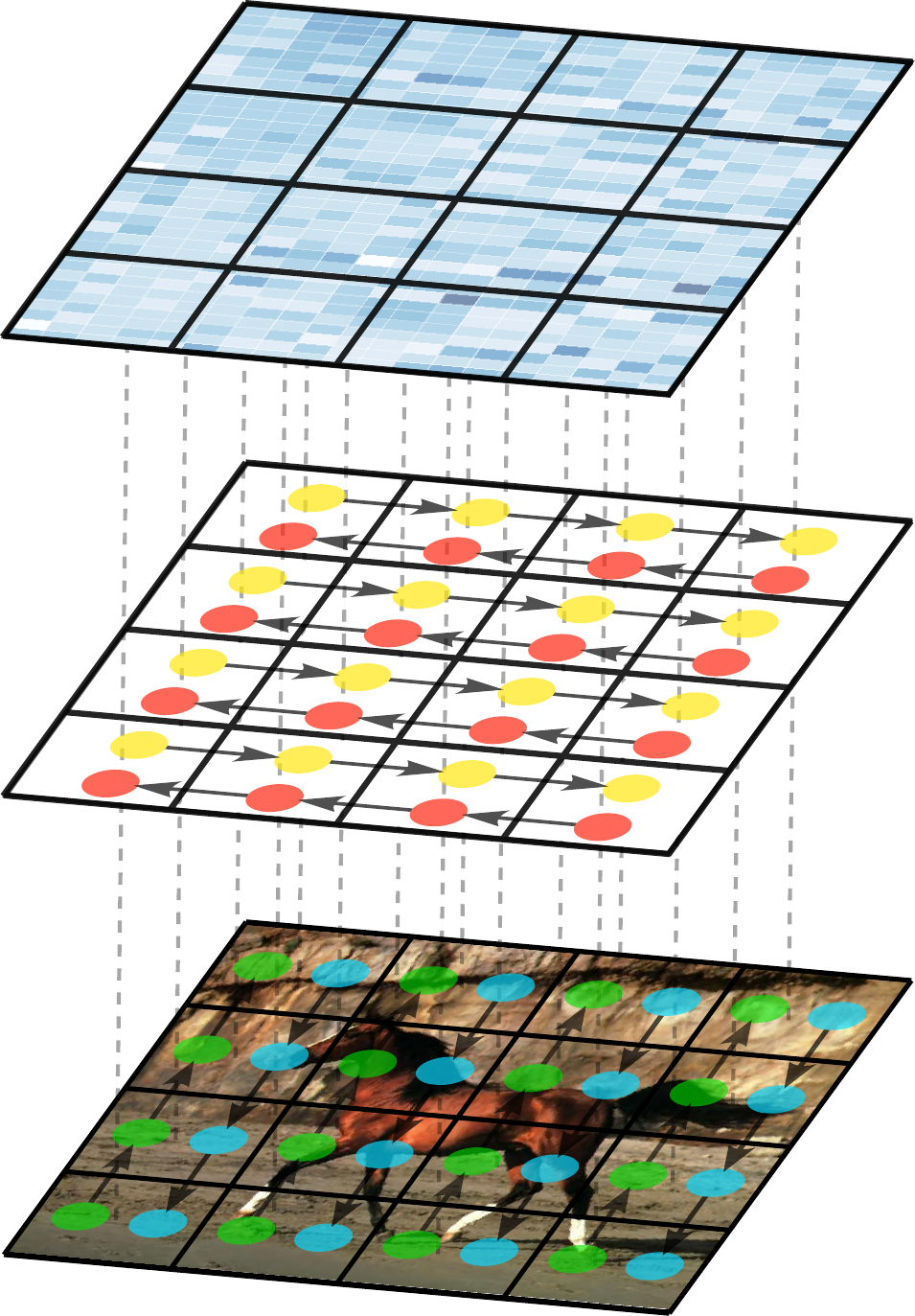}
        \caption{A ReNet layer. The blue and green dots on the input
            image/feature map represent the steps of $f^{\downarrow}$ and
            $f^{\uparrow}$ respectively. On the concatenation of the resulting
            feature maps, $f^{\rightarrow}$ (yellow dots) and $f^{\leftarrow}$
            (red dots) are subsequently swept. Their feature maps are finally
            concatenated to form the output of the ReNet layer, depicted as a
            blue heatmap in the figure.}
        \label{fig:first_layer}
        \vspace{-5mm}
    \end{center}
\end{figure}

As depicted in~\autoref{fig:first_layer}, each recurrent layer is composed
by 4 RNNs coupled together in such a way to capture the local and global
spatial structure of the input data.

%

Specifically, we take as an input an image (or the feature map of the previous
layer) $\mathbf{X}$ of elements $x \in \RR^{H \times W \times C}$, where $H$,
$W$ and $C$ are respectively the height, width and number of channels (or
features) and we split it into $I \times J$ patches $p_{i,j} \in \RR^{H_p
\times W_p \times C}$. We then sweep 
vertically a first time with two RNNs $f^{\downarrow}$ and $f^{\uparrow}$,
with $U$ recurrent units each, that move top-down and bottom-up respectively.
Note that the processing of each column is independent and can be done in
parallel.

\begin{figure*}[t]
    \advance\leftskip-0.2\textwidth
    \centering
    \makebox[\textwidth][c]{\includegraphics
    	[height=.135\textheight,width=\textwidth]{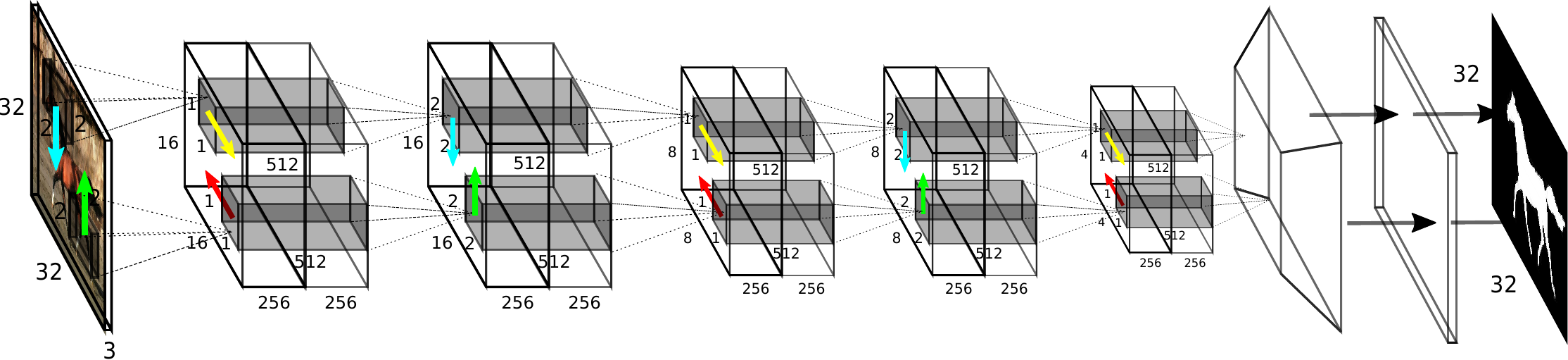}}
    \caption{The ReSeg network. For space reasons we do not represent the
        pretrained VGG-16 convolutional layers that we use to preprocess
        the input to ReSeg. The first 2 RNNs (blue and green) are applied on
        2x2x3 patches of the image, their 16x16x256 feature maps are
        concatenated and fed as input to the next two RNNs (red and yellow)
        which read 1x1x512 patches and emit the output of the first ReNet
        layer. Two similar ReNet layers are stacked, followed by an upsampling
        layer and a softmax nonlinearity.}
    \label{fig:ReSeg}
\end{figure*}

At every time step each RNN reads the next non-overlapping patch
$p_{i,j}$ and, based on its previous
state, emits a projection $o_{i,j}^{\star}$ and updates its state
$z_{i,j}^{\star}$:
\begin{align}
    o^{\downarrow}_{i,j} = f^{\downarrow}(z^{\downarrow}_{i-1,j},p_{i,j}),
        &\text{ for }i=1,\cdots, I\\
    o^{\uparrow}_{i,j} = f^{\uparrow}(z^{\uparrow}_{i+1,j},p_{i,j}),
        &\text{ for }i=I,\cdots,1
\end{align}
We stress that the decision to read non-overlapping patches is a modeling
choice to increase the image scan speed and lower the memory usage, but is not
a limitation of the architecture.

Once the first two vertical RNNs have processed the whole input $X$, we
concatenate their projections $o^{\downarrow}_{i,j}$ and $o^{\uparrow}_{i,j}$
to obtain a composite feature map $\mathbf{O^{\updownarrow}}$ whose elements
$o^{\updownarrow}_{i,j} \in \RR^{2U}$ can be seen as the activation of a
feature detector at the location $(i,j)$ with respect to all the patches in the
$j$-th column of the input. We denote what we described so far as the
\emph{vertical recurrent sublayer}.

After obtaining the concatenated feature map $\mathbf{O^{\updownarrow}}$, we
sweep over each of its rows with a pair of new RNNs, $f^{\rightarrow}$ and
$f^{\leftarrow}$. We chose not to split $\mathbf{O^{\updownarrow}}$ into
patches so that the second recurrent sublayer has the same granularity as the
first one, but this is not a constraint of the model and different
architectures can be explored.  With a similar but specular procedure as the
one described before, we proceed reading one element $o^\updownarrow_{i,j}$ at
each step, to obtain a concatenated feature map
$\mathbf{O^\leftrightarrow}~=~\left\{ h^\leftrightarrow_{i,j} \right\}
_{i=1\dots I}^{j=1\dots J}$, once again with $o^\leftrightarrow_{i,j} \in
\RR^{2U}$. Each element $o^\leftrightarrow_{i,j}$ of this \emph{horizontal
recurrent sublayer} represents the features of one of the input image patches
$p_{i,j}$ \emph{with contextual information from the whole image}.

It is trivial to note that it is possible to concatenate many recurrent layers
$\mathbf{O^{(1 \cdots L)}}$ one after the other and train them with any
optimization algorithm that performs gradient descent, as the composite model
is a smooth, continuous function.

\subsection{Upsampling layer}\label{sec:upsampling}
Since by design each recurrent layer processes non-overlapping patches, the
size of the last composite feature map will be smaller than the size of the
initial input $\mathbf{X}$, whenever the patch size is greater than one. To be
able to compute a segmentation mask at the same resolution as the ground truth,
the prediction should be expanded back before applying the softmax non-linearity.

Several different methods can be used to this end, e.g., fully connected
layers, full convolutions and transposed convolutions. The first is not a good
candidate in this domain as it does not take into account the topology of the
input, which is essential for this task; the second is not optimal either, as
it would require large kernels and stride sizes to upsample by the
required factor. Transposed convolutions are both memory and
computation efficient, and are the ideal method to tackle this problem.

Transposed convolutions -- also known as \emph{fractionally strided
convolutions} -- have been employed in many works in recent
literature~\cite{Zeiler-ICCV2011,ZeilerFergus14,long2015fully,
radford2015unsupervised,im2016generating}. This method is based on the observation
that direct convolutions can be expressed as a dot product between the
flattened input and a sparse matrix, whose non-zero elements are elements of the
convolutional kernel. The equivalence with the convolution is granted by the
connectivity pattern defined by the matrix.

Transposed convolutions apply the transpose of this transformation matrix to
the input, resulting in an operation whose input and output shapes are inverted
with respect to the original direct convolution. A very efficient implementation of
this operation can be obtained exploiting the gradient operation of the
convolution -- whose optimized implementation can be found in many of the most
popular libraries for neural networks. For an in-depth and comprehensive
analysis of each alternative, we refer the interested reader
to~\cite{dumoulin2016guide}.

\section{Experiments}\label{sec:Experiments}

\subsection{Datasets}
We evaluated the proposed ReSeg architecture on several benchmark datasets.
We proceeded by first assessing the performances of the model on the Weizmann
Horse and the Oxford Flowers datasets and then focused on the more challenging
Camvid dataset. We will describe each dataset in detail in this section.

\subsubsection{Weizmann Horse}
The Weizmann Horse dataset, introduced in~\cite{Borenstein04combiningtop-down},
is an image
segmentation dataset consisting of 329 variable size images in both RGB and
gray scale
format, matched with an equal number of groundtruth segmentation images, of
the same size as the corresponding image.
The groundtruth segmentations contain a foreground/background mask of the
focused horse, encoded as a real-value between 0 and 255. To convert this into
a boolean mask, we threshold in the center of the range setting all smaller
values to 0, and all greater values to 1.


\subsubsection{Oxford Flowers 17}
The Oxford Flowers 17 class dataset from~\cite{Nilsback06} contains 1363
variable size RGB images, with 848 image segmentations maps associated with
a subset of
the RGB images. There are 8 unique segmentation classes defined over all maps,
including flower, sky, and grass. To build a foreground/background mask,
we take the original segmentation maps, and set any pixel not belonging to
class 38 (flower class) to 0, and setting the flower class pixels to 1.
This binary segmentation task for Oxford Flowers 17 is further described
in~\cite{Xiaomeng14}.

\subsubsection{CamVid Dataset}
The Cambridge-driving Labeled Video Database
(CamVid)~\cite{Brostow2010semantic} is a real-world dataset which consists of
images recorded from a car with an internally mounted camera, capturing frames
of $960 \times 720$ RGB pixels per frame, with a recording frame rate of 30
frames per second. A total of ten minutes of video was recorded, and
approximately one frame per second has been manually annotated with per pixel
class labels, from one of 32 possible classes.  A small number of pixels were
labelled as void in the original dataset. These do not belong to any of the 32
classes prescribed in the original data, and are ignored during evaluation.  We
used the same subset of 11 class categories as~\cite{badrinarayanan2015segnet}
for experimental analysis.  The CamVid dataset itself is split into 367
training, 101 validation and 233 test images, and in order to make our
experimental setup fully comparable to~\cite{badrinarayanan2015segnet}, we
downsampled all the images by a factor of 2 resulting in a final $480 \times
360$ resolution.

\begin{table*}[!ht]
    \begin{minipage}{0.45\textwidth}
        \centering
        \small{%
            \begin{tabular}{c|c|c||c|c|c}
                \multicolumn{1}{c}{Method} & \multicolumn{1}{c}{Global acc} & \multicolumn{1}{c}{\textbf{Avg IoU}}\\ \hline \hline

                All foreground baseline & 25.4 & 79.9 \\ \hline
                All background baseline & 74.7 & 0.0 \\ \hline
                Kernelized structural SVM \cite{bertelli2011kernelized} & 94.6 & 80.1 \\ \hline
                ReSeg (no VGG) & 94.9 & 79.9 \\ \hline
                CRF learning \cite{liu2015crf} & 95.7 & 84.0 \\ \hline
                PatchCut \cite{yang2015patchcut} & 95.8 & 84.0 \\ \hline
                \textbf{ReSeg} & 96.8 & \textbf{91.6} \\ \hline

            \end{tabular}
            \vspace*{0.1cm}
        }
        \caption{Weizmann Horses. Per pixel accuracy and IoU are
            reported.}
        \label{tbl:WeizmannHorses_SOTA}
    \end{minipage}
    \quad
    \begin{minipage}{0.45\textwidth}
        \centering
        \small{%
            \begin{tabular}{c|c|c||c|c|c}
                \multicolumn{1}{c}{Method} & \multicolumn{1}{c}{Global acc} & \multicolumn{1}{c}{\textbf{Avg IoU}}\\ \hline \hline

                All background baseline & 71.0 & 0.0 \\ \hline
                All foreground baseline & 29.0 & 29.2 \\ \hline
                GrabCut \cite{rother2004grabcut} & 95.9 & 89.3 \\ \hline
                Tri-map \cite{Xiaomeng14} & 96.7 & 91.7 \\ \hline
                \textbf{ReSeg} & 98 & \textbf{93.7} \\ \hline

            \end{tabular}
            \vspace*{0.1cm}
        }
        \caption{Oxford Flowers. Per pixel accuracy and IoU are
            reported.}
        \label{tbl:OxfordFlowers_SOTA}
    \end{minipage}
\end{table*}

\begin{table*}[!ht]
	\resizebox{\textwidth}{!}{%
		\small{%
            \begin{tabular}{c|c|c|c|c|c|c|c|c|c|c|c||c|c|c}

                \multicolumn{1}{c}{Method} & \multicolumn{1}{c}{\rotatebox{90}{Building}} & \multicolumn{1}{c}{\rotatebox{90}{Tree}} & \multicolumn{1}{c}{\rotatebox{90}{Sky}} & \multicolumn{1}{c}{\rotatebox{90}{Car}} & \multicolumn{1}{c}{\rotatebox{90}{Sign-Symbol}} & \multicolumn{1}{c}{\rotatebox{90}{Road}} & \multicolumn{1}{c}{\rotatebox{90}{Pedestrian}} & \multicolumn{1}{c}{\rotatebox{90}{Fence}} & \multicolumn{1}{c}{\rotatebox{90}{Column-Pole}} & \multicolumn{1}{c}{\rotatebox{90}{Side-walk}} & \multicolumn{1}{c}{\rotatebox{90}{Bicyclist}} & \multicolumn{1}{c}{\rotatebox{90}{Avg class acc}} & \multicolumn{1}{c}{\rotatebox{90}{Global acc}} & \multicolumn{1}{c}{\rotatebox{90}{\textbf{Avg IoU}}}\\ \hline \hline

				\multicolumn{15}{c}{\emph{Segmentation models}} \\ \hline
                Super Parsing   \cite{tighe2013superparsing} & \textbf{87.0} & 67.1 & \textbf{96.9} & 62.7 & 30.1 & 95.9 & 14.7 & 17.9 & 1.7 & 70.0 & 19.4 & 51.2 & 83.3 & n/a \\ \hline
				Boosting+Higher order \cite{sturgess2009combining} & 84.5 & 72.6 & \textbf{97.5} & 72.7 & 34.1 & 95.3 & 34.2 & 45.7 & 8.1 & 77.6 & 28.5 & 59.2 & 83.8 & n/a \\ \hline
				Boosting+Detectors+CRF \cite{ladicky2010and} & 81.5 & 76.6 & 96.2 & 78.7 & 40.2 & 93.9 & 43.0 & 47.6 & 14.3 & 81.5 & 33.9 & 62.5 & 83.8 & n/a \\ \hline

				\multicolumn{15}{c}{\emph{Neural Network based segmentation models}} \\ \hline
				SegNet-Basic (layer-wise training \cite{badrinarayanan2015segnetlayerwise}) & 75.0 & 84.6 & 91.2 & 82.7 & 36.9 & 93.3 & 55.0 & 37.5 & 44.8 & 74.1 & 16.0 & 62.9 & 84.3 & n/a \\ \hline
                SegNet-Basic \cite{badrinarayanan2015segnet} & 80.6 & 72.0 & 93.0 & 78.5 & 21.0 & 94.0 & 62.5 & 31.4 & 36.6 & 74.0 & 42.5 & 62.3 & 82.8 & 46.3 \\ \hline
                SegNet \cite{badrinarayanan2015segnet} & \textbf{88.0 } & \textbf{87.3} & 92.3 & 80.0 & 29.5 & \textbf{97.6} & 57.2 & \textbf{49.4} & 27.8 & 84.8 & 30.7 & 65.9 & 88.6 & 50.2 \\ \hline
                \emph{ReSeg + Class Balance} & 70.6 & 84.6 & 89.6 & 81.1 & \textbf{61.0} & 95.1 & \textbf{80.4} & 35.6 & \textbf{60.6} & \textbf{86.3} & \textbf{60.0} & 73.2 & 83.5 & 53.7 \\ \hline
                \textbf{ReSeg} & 86.8 & 84.7 & 93.0 & \textbf{87.3} & 48.6 & \textbf{98.0} & 63.3 & 20.9 & 35.6 & \textbf{87.3} & 43.5 & 68.1 & 88.7 & \textbf{58.8} \\ \hline

				\multicolumn{15}{c}{\emph{Sub-model averaging}} \\ \hline
                \emph{Bayesian SegNet-Basic} \cite{Kendall2015bayesiansegnet} & 75.1 & 68.8 & 91.4 & 77.7 & 52.0 & 92.5 & 71.5 & 44.9 & 52.9 & 79.1 & 69.6 & 70.5 & 81.6 & 55.8 \\ \hline
                \emph{Bayesian SegNet} \cite{Kendall2015bayesiansegnet} & 80.4 & 85.5 & 90.1 & 86.4 & 67.9 & 93.8 & 73.8 & 64.5 & 50.8 & 91.7 & 54.6 & 76.3 & 86.9 & 63.1 \\ \hline

			\end{tabular}
		}
    }
	\vspace*{0.1cm}
    \caption{%
        CamVid. The table reports the per-class accuracy, the average per-class
        accuracy, the global accuracy and the average intersection over union.
        The best values and the values within $1$ point from the best are
        highlighted in bold for each column. For completeness we report the
        Bayesian Segnet models even if they are not directly comparable to the
        others as they perform a form of model averaging.}
    \label{tbl:camvid_SOTA}
\end{table*}

\begin{table*}[!ht]
    \resizebox{\textwidth}{!}{%
        \small{%
            \begin{tabular}{c|c|c|c|c|c|c|c|c|c|c|c|c|c|c|c||c|c|c}
                \multicolumn{1}{c}{Model} & \multicolumn{1}{c}{${ps}_{\text{RE}}$}& \multicolumn{1}{c}{$d_{\text{RE}}$} & \multicolumn{1}{c}{${fs}_{\text{UP}}$}& \multicolumn{1}{c}{$d_{\text{UP}}$} & \multicolumn{1}{c}{\rotatebox{90}{Building}} & \multicolumn{1}{c}{\rotatebox{90}{Tree}} & \multicolumn{1}{c}{\rotatebox{90}{Sky}}  & \multicolumn{1}{c}{\rotatebox{90}{Car}}  & \multicolumn{1}{c}{\rotatebox{90}{Sign-Symbol}} & \multicolumn{1}{c}{\rotatebox{90}{Road}} & \multicolumn{1}{c}{\rotatebox{90}{Pedestrian}} & \multicolumn{1}{c}{\rotatebox{90}{Fence}} & \multicolumn{1}{c}{\rotatebox{90}{Column-Pole}} & \multicolumn{1}{c}{\rotatebox{90}{Side-walk}} & \multicolumn{1}{c}{\rotatebox{90}{Bicyclist}} & \multicolumn{1}{c}{\rotatebox{90}{Avg class acc}} & \multicolumn{1}{c}{\rotatebox{90}{Global acc}} & \multicolumn{1}{c}{\rotatebox{90}{\textbf{Avg IoU}}}\\ \hline \hline

                ReSeg + LCN & $(2 \times 2), (1 \times 1)$ & (100, 100) & $(2 \times 2)$ & (50, 50) & 81.5 & 80.3 & \textbf{94.7} & 78.1 & 42.8 & \textbf{97.4} & 53.5 & 34.3 & 36.8 & 68.9 & 47.9 & 65.1 & 84.8 & 52.6 \\ \hline
                ReSeg + Class Balance & $(2 \times 2), (1 \times 1)$ & (100, 100) & $(2 \times 2)$ & (50, 50) & 70.6 & \textbf{84.6} & 89.6 & 81.1 & \textbf{61.0} & 95.1 & \textbf{80.4} & \textbf{35.6} & \textbf{60.6} & \textbf{86.3} & \textbf{60.0} & 73.2 & 83.5 & 53.7 \\ \hline
                \textbf{ReSeg} & $(2 \times 2), (1 \times 1)$ & (100, 100) & $(2 \times 2)$ & (50, 50) & \textbf{86.8} & \textbf{84.7} & 93.0 & \textbf{87.3} & 48.6 & \textbf{98.0} & 63.3 & 20.9 & 35.6 & \textbf{87.3} & 43.5 & 68.1 & 88.7 & \textbf{58.8} \\ \hline
            \end{tabular}%
        }
    }
    \vspace*{0.1cm}
    \caption{%
        Comparison of the performance of different hyperparameter on CamVid.
    }
    \label{tbl:camvid_params}
\end{table*}
\subsection{Experimental settings}
To gain confidence with the sensitivity of the model to the different
hyperparameters, we decided to evaluate it first on the Weissman Horse and
Oxford Flowers datasets on a binary segmentation task; we then focused the most
of our efforts on the more challenging semantic segmentation task on the CamVid
dataset.

The number of hyperparameters of this model is potentially very high, as for
each ReNet layer different implementations are possible (namely vanilla RNN,
GRU or LSTM), each one with its specific parameters. Furthermore, the number of
features, the size of the patches and the initialization scheme have to be
defined for each ReNet layer as well as for each transposed convolutional
layer. To make it feasible to explore the hyperparameter space, some of the
hyperparameters have been fixed by design and the remaining have been finetuned.
In the rest of this section, the architectural choices for both sets of
parameters will be detailed.

All the transposed convolution upsampling layers were followed by a
ReLU~\cite{Krizhevsky2012-alexnet} non-linearity and initialized with the
fan-in plus fan-out initialization scheme described
in~\cite{glorot2010understanding}. The recurrent weight matrices were instead
initialized to be orthonormal, following the procedure defined
in~\cite{Saxe2014}. We also constrained the stride of the upsampling transposed
convolutional layers to be tied to their filter size.

In the segmentation task, each training image carries classification
information for all of its pixels. Differently from the image classification
task, small batch sizes provide the model with a good amount of information
with sufficient variance to learn and generalize well. We experimented with
various batch sizes going as low as processing a single image at the time,
obtaining comparable results in terms of performance. In our experiments we
kept a fixed batch size of $5$, as a compromise between train speed and memory
usage. In all our experiments, we used L2
regularization~\cite{Krogh92asimple}, also known as weight decay, set to
$0.001$ to avoid instability at the end of training. We trained all our models
with the Adadelta~\cite{Zeiler-2012} optimization algorithm, for its desired
property of not requiring a specific hyperparameter tuning. The effect of Batch
Normalization in RNNs has been a focus of attention~\cite{Laurent2015}, but it
does not seem to provide a reliable improvement in performance, so we decided
not to adopt it.

In the experiments, we varied the number of ReNet layers and the number of
upsampling transposed convolutional layers, each of them defined respectively
by the number of features $d_{\text{RE}}(l)$ and $d_{\text{UP}}(l)$,
the size of the input patches (or equivalently of the filters)
${ps}_{\text{RE}}(l)$ and ${fs}_{\text{UP}}(l)$.

\subsection{Results}

In~\autoref{tbl:WeizmannHorses_SOTA}, we report the results on the Weizmann
Horse dataset. On this dataset, we verified the assumption that processing
the input image with some pre-trained convolutional layers from VGG-16 could
ease the learning. Specifically, we restricted ourselves to only using the
first $7$ convolutional layers from VGG, as we only intended to extract some
low-level generic features and learn the task-specific high-level features with
the ReNet layers. The results indeed show an increase in terms of average Intersection
over Union (\emph{IoU}) when these layers are being used, confirming our
hypothesis.

\autoref{tbl:OxfordFlowers_SOTA} shows the results for Oxford Flowers
dataset, when using the full ReSeg architecture (i.e., including VGG convolutional layers).
As shown in the table, our method clearly outperforms the
state-of-the-art both in terms of global accuracy and average IoU.

\autoref{tbl:camvid_SOTA} presents the results on CamVid dataset using the full
ReSeg architecture. Our model exhibits state-of-the-art performance in terms of
IoU when compared to both standard segmentation methods and neural network
based methods, showing an increase of $17\%$ w.r.t.\ to the recent SegNet
model. It is worth highlighting that incorporating sub-model averaging to
SegNet model, as in \cite{Kendall2015bayesiansegnet}, boosts the original model
performance, as expected. Therefore, introducing sub-model averaging to ReSeg
would also presumably result in significant performance increase.  However,
this remains to be tested.




\section{Discussion}

As reported in the previous section, our experiments on the Weizmann Horse
dataset show that processing the input images with some layers of VGG-16
pre-trained network improves the results. In this setting, pre-processing the input
with Local Contrast Normalization (LCN) does not seem to give any advantage
(see~\autoref{tbl:camvid_params}). We did not use any other kind of
pre-processing.

\begin{figure*}[!htb]
    \centering
    \includegraphics[width=\textwidth]{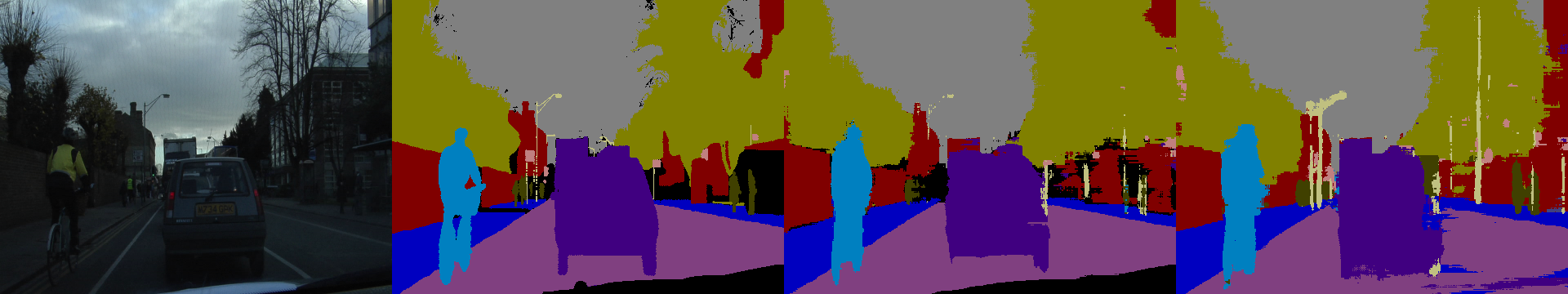}
    \caption{Camvid segmentation example with and without class balancing. From
        the left: input image, ground truth segmentation, ReSeg segmentation,
        ReSeg segmentation with class balancing. Class balancing improves the
        low frequency classes as e.g., the street lights, at the price of a
        worse overall segmentation.}
    \label{fig:camvid_class_balance}
\end{figure*}

While on both the Weizmann Horse and the Oxford Flowers datasets we trained on
a binary background/foreground segmentation task, on CamVid we addressed the
full semantic segmentation task. In this setting, when the dataset is highly
imbalanced, the segmentation performance of some classes can drop
significantly as the network tries to maximize the score on the high-occurrence
classes, {\em de facto} ignoring the low-occurrence ones. To overcome this behaviour,
we added a term to the cross-entropy loss to bias the prediction towards the
low-occurrence classes. We use \emph{median frequency
balancing}~\cite{Eigen2015}, which re-weights the class predictions by the
ratio between the median of the frequencies of the classes (computed on the
training set) and the frequency of each class.
This increases the score of the low frequency classes
(see~\autoref{tbl:camvid_params}) at the price of a more noisy segmentation
mask, as the probability of the underrepresented classes is overestimated and
can lead to an increase in misclassified pixels in the output segmentation
mask, as shown in~\autoref{fig:camvid_class_balance}.

On all datasets we report the per-pixel accuracy (\emph{Global acc}), computed
as the percentage of true positives w.r.t.\ the total number of pixels in the
image, and the average per-class Intersection over Union (\emph{Avg IoU}),
computed on each class as true positive divided by the sum of true positives,
false positives and false negatives and then averaged. In the full semantic
segmentation setting we also report the per-class accuracy and the average
per-class accuracy (\emph{Avg class acc}).

\section{Conclusion}
We introduced the ReSeg model, an extension of the ReNet model for image
semantic segmentation. The proposed architecture shows state-of-the-art
performances on CamVid, a widely used dataset for urban scene semantic
segmentation, as well as on the much smaller Oxford Flowers dataset. We also
report state-of-the-art performances on the Weizmann Horses.

In our analysis, we discuss the effects of applying some layers of VGG-16 to
process the input data, as well as those of introducing a class balancing
term in the cross-entropy loss function to help the learning of
under-represented classes.
Notably, it is sufficient to process the input images with just a few layers of
VGG-16 for the ReSeg model to gracefully handle the semantic segmentation task, confirming
its ability to encode contextual information and long term dependencies.

\subsubsection*{Acknowledgments}
We would like to thank all the developers of
Theano~\cite{Bergstra2010,Bastien2012} and in particular Pascal Lamblin, Arnaud
Bergeron and Frédéric Bastien for their dedication. We are also thankful to
César Laurent for the moral support and to Vincent Dumoulin for the insightful
discussion on transposed convolutions. We are also very grateful to the
developers of Lasagne~\cite{Lasagne} for providing a light yet powerful
framework and to the reviewers for their valuable feedback. We finally
acknowledge the support of the following organizations for research funding and
computing support: NSERC, IBM Watson Group, IBM Research, NVIDIA, Samsung,
Calcul Qu\'{e}bec, Compute Canada, the Canada Research Chairs and CIFAR. F.V.
was funded by the AI*IA Young Researchers Mobility Grant and the Politecnico di
Milano PHD School International Mobility
Grant.

{\small
\bibliographystyle{ieee}
\bibliography{myrefs,ml,segm,camvid}
}

\end{document}